\pdfoutput=1
\documentclass[11pt]{article}

\PassOptionsToPackage{table}{xcolor}

\usepackage[preprint]{acl}

\usepackage{times}
\usepackage{latexsym}
\usepackage{xcolor}
\usepackage{colortbl}

\usepackage[T1]{fontenc}
\usepackage[utf8]{inputenc}
\usepackage{microtype}

\usepackage{graphicx}
\usepackage{url}
\usepackage{multirow}
\usepackage{booktabs}
\usepackage{array}
\usepackage{subcaption}
\usepackage{amsmath}
\usepackage{amsfonts}
\usepackage{amssymb}
\usepackage{bm}
\usepackage{caption}
\usepackage{adjustbox}
\usepackage{tabularx}
\usepackage{xspace}
\usepackage{cleveref}
\usepackage{tikz}
\usepackage{etoc}
\usepackage{tcolorbox}
\usepackage{dblfloatfix}
\usepackage{todonotes}
\usepackage{natbib}

\definecolor{lightgray}{gray}{0.9}
\definecolor{darkgray}{gray}{0.7}
\definecolor{burntorange}{RGB}{204, 85, 0}
\definecolor{darkgreen}{rgb}{0.0, 0.5, 0.0}

\title{Decomposing Prompts into Reusable Scaffolds and Task-Specific Residuals}

\author{
    Yupeng Chang$^{1}$ \quad Yi Chang$^{1,2,3}$ \quad Yuan Wu$^{1}$\footnotemark[1] \\
    $^{1}$School of Artificial Intelligence, Jilin University \\
    $^{2}$Engineering Research Center of Knowledge-Driven Human-Machine Intelligence, MOE, China \\
    $^{3}$International Center of Future Science, Jilin University \\
    \texttt{changyp23@mails.jlu.edu.cn}, \texttt{\{yichang, yuanwu\}@jlu.edu.cn}
}

\begin{document}

\etocdepthtag.toc{chapter}
\etocsettagdepth{chapter}{none}
\etocsettagdepth{appendix}{none}

\maketitle
\renewcommand{\thefootnote}{\fnsymbol{footnote}}
\footnotetext[1]{Corresponding author}

\begin{abstract}
Automatic prompt optimization typically treats each prompt as a monolithic text string, requiring task-specific search from scratch and limiting reuse across related tasks. We study a more structured view in which a prompt is decomposed into two components: a reusable \textbf{Instructional Scaffold} that captures shared domain-level instructions, and a concise \textbf{Task-specific Residual} that introduces task-level refinements. Based on this view, we propose \textbf{DiSR} (\textbf{D}iscovering \textbf{i}nstructional \textbf{S}caffolds and \textbf{R}esiduals), a two-stage framework motivated by the intuition that related tasks may share a compact instructional core while differing in a small amount of task-specific information. DiSR first discovers a scaffold from source tasks within a coherent domain and then adapts it to a target task by learning a residual. Across multiple benchmarks, DiSR provides a stronger starting point than manual instructions and monolithic prompt optimization, yielding competitive or improved task performance and often better-calibrated confidence estimates. Additional analyses examine scaffold generalization, residual sparsity, and cross-domain portability, supporting the view that prompt reuse can be studied through scaffold--residual decomposition. Overall, the results suggest a practical and analyzable alternative to optimizing each prompt independently.
\end{abstract}

\section{Introduction}

Large language models (LLMs) are highly sensitive to prompt design, and recent automatic prompt optimization (APO) methods have shown that model-generated instructions can outperform manually written prompts on many tasks. Representative approaches include LLM-as-optimizer methods such as OPRO~\citep{yang2024largelanguagemodelsoptimizers} and related methods~\citep{zhou2022large,pryzant2023automatic,wang2023promptagent}, as well as gradient-based approaches such as AutoPrompt~\citep{shin2020autoprompt}. Despite their differences, these methods typically optimize a prompt for each task independently, treating the prompt as a single text object to be rewritten or searched from scratch.

This monolithic view is effective when tasks are considered in isolation, but it makes reuse across related tasks difficult. In many NLP settings, tasks within the same task family share substantial instructional commonality: they often involve similar output constraints, similar styles of reasoning, and similar domain-specific role framing. Yet standard APO pipelines do not explicitly model such shared structure. As a result, instruction patterns discovered for one task are not clearly separated from task-specific details, which limits reuse and makes it harder to analyze what transfers across tasks.

In this work, we study whether prompts within a coherent task domain can be decomposed into a reusable component and a task-specific component. Our focus differs from work on decomposing complex tasks into multi-step workflows~\citep{khot2022decomposed}. Instead, we study the \emph{internal structure of a single prompt}. We hypothesize that prompts for related tasks can often be represented as a shared domain-level instructional scaffold together with a relatively small task-level refinement. This view is also consistent with an MDL-inspired intuition: if related tasks share regularities, it may be better to capture those regularities once and express task differences through compact edits rather than re-optimizing an entire prompt from scratch.

We instantiate this view with two components. An \textbf{Instructional Scaffold} is a reusable prompt backbone that captures broad domain-level instructions, such as role framing, expected answer format, and general reasoning style. A \textbf{Task-specific Residual} is a concise adaptation that specializes the scaffold for a particular task. For example, within a clinical domain, a scaffold may specify the role, decision format, and general diagnostic framing, while different residuals adapt that scaffold to tasks centered on patient history, laboratory findings, or treatment selection.

Based on this decomposition, we propose \textbf{DiSR} (\textbf{D}iscovering \textbf{i}nstructional \textbf{S}caffolds and \textbf{R}esiduals), a two-stage framework for automatic prompt optimization. In the first stage, \emph{Scaffold Discovery}, DiSR identifies a reusable scaffold from a set of source tasks within a domain. In the second stage, \emph{Residual Learning}, it adapts the scaffold to a target task by learning a concise residual rather than re-optimizing the entire prompt from scratch. This design allows us to study prompt reuse in a controlled setting while keeping the underlying optimizer fixed.

Our contributions are as follows:
\begin{itemize}
    \item We formulate prompt optimization as a scaffold--residual decomposition problem, providing an explicit framework for studying reusable instruction structure across related tasks.
    \item We introduce \textbf{DiSR}, a two-stage framework that first discovers a domain-level scaffold and then adapts it to a target task through a task-specific residual.
  \item We show empirically that this decomposition can improve prompt reuse and target-task adaptation, yielding strong performance and often better-calibrated confidence estimates relative to monolithic baselines.
    \item We present analyses of scaffold generalization, residual sparsity, and portability to clarify what transfers across tasks and where the decomposition is most effective.
\end{itemize}

\noindent\textbf{Code repository:} {\small\url{https://github.com/llm172/DiSR}}.

\section{Related Work} \label{sec:related_work}

Our work is most closely related to Automated Prompt Optimization (APO), which aims to improve prompt design without relying on manual trial and error. Existing APO methods mainly fall into two categories: methods that use LLMs as optimizers to iteratively propose and refine prompts, such as OPRO and Automatic Prompt Engineer~\citep{yang2024largelanguagemodelsoptimizers,zhou2022large}, and gradient-based token search methods such as AutoPrompt~\citep{shin2020autoprompt}. Despite their differences, these approaches typically optimize a separate prompt for each task and treat the prompt as a single text object. DiSR instead asks whether prompts for related tasks can be decomposed into a shared instructional component and a task-specific adaptation. In this sense, our framework is complementary to existing APO methods rather than a replacement for them: we keep the underlying optimizer fixed and study whether an explicit scaffold--residual decomposition improves reuse and adaptation across related tasks.

Our work is also related to modular and structured prompting, but it differs from methods such as Decomposed Prompting~\citep{khot2022decomposed}, which decompose a complex task into multiple sub-prompts or intermediate reasoning steps at the workflow level. By contrast, DiSR studies the internal structure of a single prompt by separating a reusable scaffold from a task-specific residual within the same prompt. It is also distinct from manual prompt curation efforts such as PromptSource~\citep{bach2022promptsource}, which organize prompts but do not automatically discover reusable prompt structure through optimization. More broadly, DiSR is loosely connected to parameter-efficient adaptation methods such as adapters and LoRA~\citep{houlsby2019parameter,hu2021lora}, in that both distinguish shared structure from relatively small task-specific updates. We use this comparison only as intuition, and likewise treat the MDL perspective~\citep{rissanen1978modeling} as a motivating lens rather than a theoretical contribution. Our contribution is an explicit and empirically testable framework for studying reusable prompt structure across related tasks.

\begin{figure*}[t]
    \centering
    \includegraphics[width=.96\textwidth]{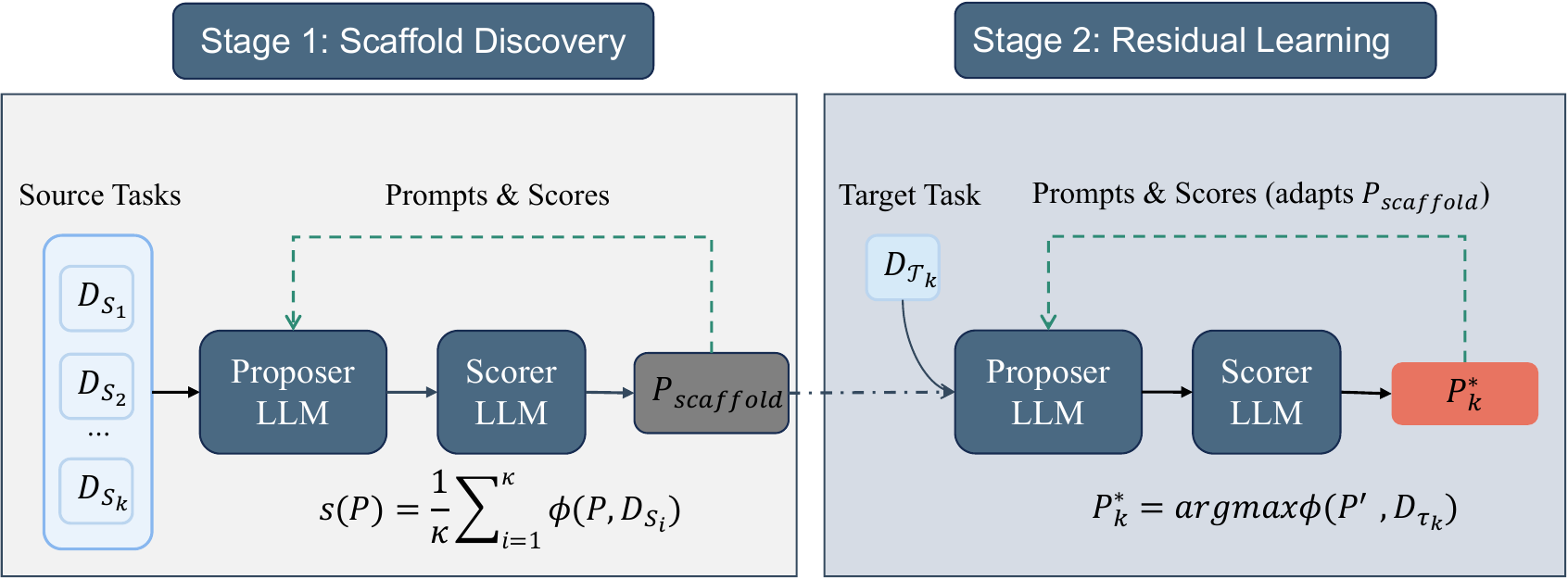}
    \caption{
        Overview of the DiSR framework.
        \textbf{Stage 1 (Scaffold Discovery):} a shared Instructional Scaffold ($P_{\text{scaffold}}$) is learned from multiple source tasks within a domain.
        \textbf{Stage 2 (Residual Learning):} the scaffold is adapted to a target task by learning a Task-specific Residual ($\Delta P_k$), producing the final prompt ($P_k^*$).
    }
    \label{fig:framework}
\end{figure*}

\section{Methodology} \label{sec:methods}

\subsection{Definition of a Task Domain}
Our framework assumes that some tasks share reusable instructional structure. To make this assumption operational, we define a \emph{task domain} as a set of tasks that share: (i) a common input/output format, (ii) the same evaluation metric, and (iii) a broadly similar reasoning or decision pattern. Examples in this work include multiple-choice question answering domains such as commonsense reasoning, medicine, law, and finance.

This definition is intended as an operational criterion for constructing source and target task sets, rather than as a universal ontology of task relatedness. In our experiments, we keep the input/output format and evaluation metric fixed within each domain while allowing source and target tasks to vary in topic or sub-type. Section~\ref{sec:experiments} further examines how discovered scaffolds generalize across held-out sub-tasks within each domain.

\subsection{Preliminaries and Problem Formulation}
For each domain, let $\mathcal{S} = \{\mathcal{S}_1, \dots, \mathcal{S}_\kappa\}$ denote a set of source tasks and $\mathcal{T} = \{\mathcal{T}_1, \dots, \mathcal{T}_\tau\}$ denote a set of target tasks. Each target task $\mathcal{T}_k$ is associated with a dataset $D_{\mathcal{T}_k} = \{(q_n, a_n)\}_{n=1}^{N_k}$, where $q_n$ is an input instance and $a_n$ is the corresponding reference output used to evaluate prompt quality.

Conventional prompt optimization learns a separate prompt $P_k^*$ for each target task $\mathcal{T}_k$ by maximizing a task-specific performance metric $\phi$:
\begin{equation} \label{eq:traditional_opt}
    P_k^* = \arg\max_{P} \phi(P, D_{\mathcal{T}_k}).
\end{equation}
This formulation treats each prompt independently and does not explicitly model shared structure across related tasks.

We instead study whether prompts for related tasks can be represented through a shared scaffold together with compact task-specific refinements. Concretely, we aim to learn a domain-level \textbf{Instructional Scaffold} $P_{\text{scaffold}}$ and, for each target task, a \textbf{Task-specific Residual} $\Delta P_k$ such that
\begin{equation}
    P_k^* = P_{\text{scaffold}} \oplus \Delta P_k,
\end{equation}
where $\oplus$ denotes composing the scaffold with a small set of task-specific edits.

This view is motivated by the following MDL-inspired objective:
\begin{equation}
\begin{aligned}
L_{\mathrm{MDL}}
&= L\big(P_{\text{scaffold}}\big)
 + \sum_{k=1}^{\tau} L\big(\Delta P_k\big) \\
&\quad + \sum_{k=1}^{\tau} L_{\mathrm{err}}\big(P_k^*, D_{\mathcal{T}_k}\big),
\end{aligned}
\label{eq:mdl_objective}
\end{equation}
where $L(\cdot)$ is a description-length proxy and $L_{\mathrm{err}}(\cdot)$ denotes task-specific error. This objective should be interpreted as motivation rather than as an exact optimization target. Intuitively, it favors representing related tasks through a shared prompt component plus compact task-specific edits while maintaining low empirical error. DiSR can be viewed as a practical two-stage approximation to this idea: Stage 1 learns a reusable scaffold, and Stage 2 adds task-specific residuals to improve target-task performance.

\subsection{The DiSR Framework}
To operationalize this decomposition, we propose \textbf{DiSR}, a two-stage framework that separates domain-level scaffold discovery from task-level adaptation. Figure~\ref{fig:framework} provides an overview.

For a target task $\mathcal{T}_k$, DiSR does not optimize a prompt from scratch. Instead, it starts from a learned scaffold $P_{\text{scaffold}}$ and adapts it using task-specific data $D_{\mathcal{T}_k}$:
\begin{equation} \label{eq:transformation}
    P_{\text{scaffold}} \xrightarrow{\text{Adapt}(D_{\mathcal{T}_k})} P_k^* .
\end{equation}
The resulting change is represented as a \textbf{Task-specific Residual} $\Delta P_k$, so that
\begin{equation}
    P_k^* = P_{\text{scaffold}} \oplus \Delta P_k.
\end{equation}

Operationally, $\Delta P_k$ should be understood as the task-specific edits introduced during Stage 2, rather than as a separately optimized free-form prompt. This distinction is important: DiSR is designed to preserve the scaffold as a reusable backbone and to specialize it through relatively small modifications, allowing us to analyze what is shared and what must change across related tasks.

\subsection{Stage 1: Scaffold Discovery}
The goal of Stage 1 is to discover an Instructional Scaffold that performs well across multiple source tasks within a domain. Optimizing a single prompt jointly over source tasks encourages the resulting scaffold to retain domain-level instruction patterns that generalize beyond any single task.

\noindent\textbf{Input:} a set of source datasets $\{D_{\mathcal{S}_1}, \dots, D_{\mathcal{S}_\kappa}\}$ and an initial generic prompt $P_0$.

\noindent\textbf{Process:} We use the same underlying optimization procedure described in Section~\ref{subsec:optimization_process}. Each candidate prompt $P$ is evaluated on all source datasets, and its score is the average performance:
\begin{equation} \label{eq:scaffold_score}
    s(P) = \frac{1}{\kappa} \sum_{i=1}^{\kappa} \phi(P, D_{\mathcal{S}_i}).
\end{equation}
Because the objective averages performance across source tasks, prompts that overfit to a single source task are disfavored, while prompts that capture shared instructional patterns are preferred.

\noindent\textbf{Output:} The highest-scoring prompt under this cross-task objective is selected as the Instructional Scaffold:
\begin{equation} \label{eq:scaffold_def}
    P_{\text{scaffold}} = \arg\max_{P} s(P).
\end{equation}

\subsection{Stage 2: Residual Learning}
Given a discovered scaffold, Stage 2 adapts it to a target task by learning a task-specific residual. The goal is not to rewrite the prompt from scratch, but to preserve the reusable domain-level structure encoded in $P_{\text{scaffold}}$ while introducing only the additional task-specific information needed for $\mathcal{T}_k$.

\noindent\textbf{Input:} the discovered scaffold $P_{\text{scaffold}}$ and a target-task dataset $D_{\mathcal{T}_k}$.

\noindent\textbf{Process:} Stage 2 reuses the same optimization engine as Stage 1, but with two changes. First, optimization is initialized from $P_{\text{scaffold}}$ rather than from the generic manual prompt $P_0$. Second, candidate prompts are evaluated only on the target dataset $D_{\mathcal{T}_k}$. The resulting objective is
\begin{equation}
P_k^* = \arg\max_{P'} \phi(P', D_{\mathcal{T}_k}).
\label{eq:residual_opt}
\end{equation}
Stage~2 optimization is initialized from $P_{\text{scaffold}}$.

To encourage concise adaptation, the Stage 2 meta-prompt explicitly conditions on $P_{\text{scaffold}}$ and instructs the Proposer LLM to preserve the scaffold while making small, task-specific edits. We refer to these edits collectively as the \textbf{Task-specific Residual} $\Delta P_k$. In this sense, $\Delta P_k$ is not a separately optimized free-form prompt, but the set of task-level modifications that transform the reusable scaffold into the final prompt $P_k^*$.

\noindent\textbf{Output:} the final task-specialized prompt $P_k^*$.

\subsection{Core Optimization Process} \label{subsec:optimization_process}
Both stages of DiSR use the same underlying optimization procedure, adapted from the LLM-as-optimizer setting of OPRO~\citep{yang2024largelanguagemodelsoptimizers}. Keeping the optimizer fixed reduces differences due to the search algorithm, allowing us to examine the effects of scaffold--residual decomposition and source-task reuse. The optimization engine consists of three components.

\noindent\textbf{Proposer LLM:} a generator model that proposes new candidate instructions at each iteration based on a meta-prompt.

\noindent\textbf{Scorer LLM:} a model that instantiates the objective function $\phi$ by executing candidate prompts on input instances and scoring the resulting outputs against reference answers.

\begin{figure*}[t!]
    \centering
    \includegraphics[width=0.96\textwidth]{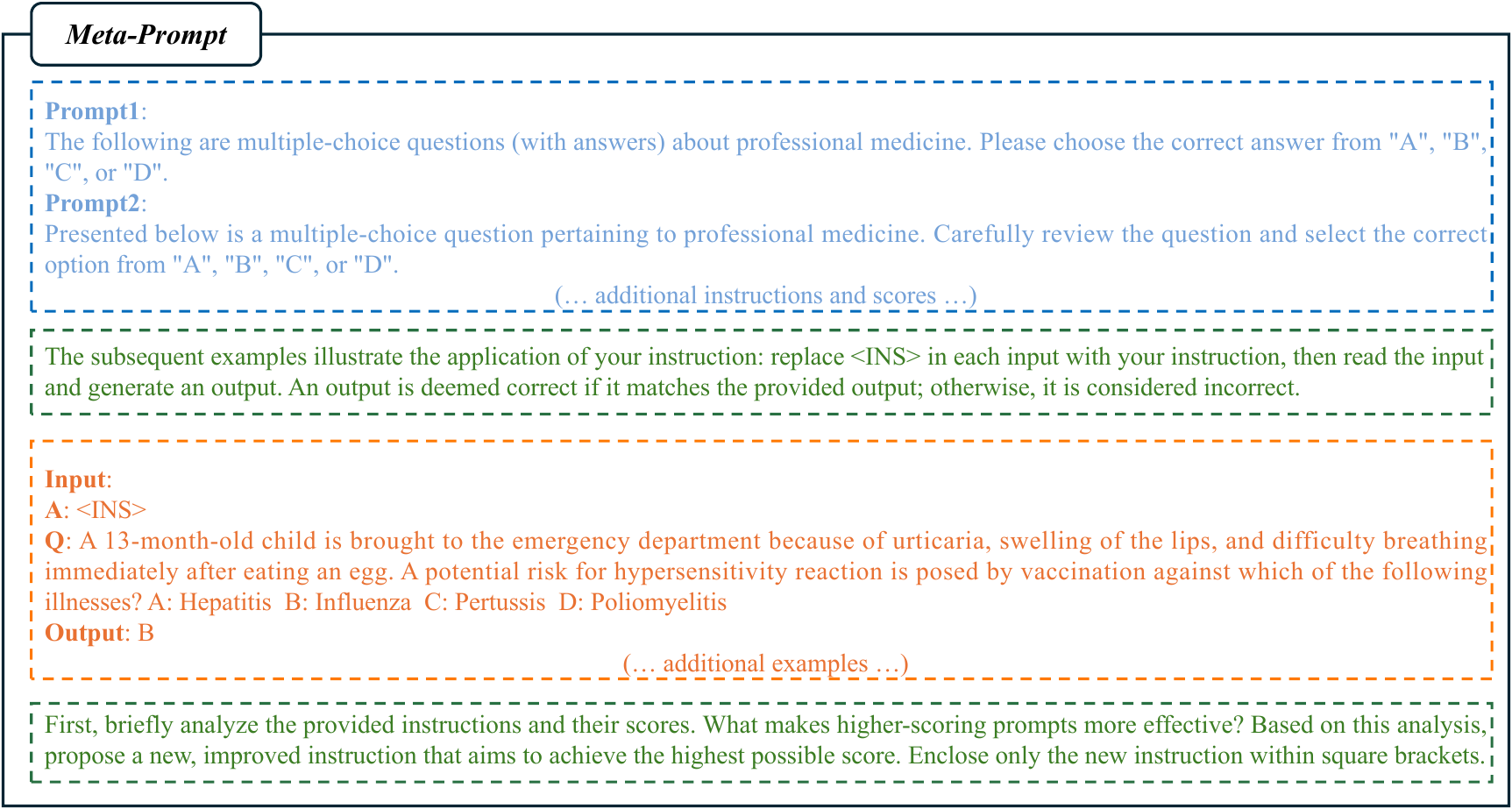}
    \caption{
        Structure of the meta-prompt used to guide the Proposer LLM: (1) \textbf{Optimization History}, a list of previous prompts and their scores; (2) \textbf{Task Schema}, an example of the expected input/output format; and (3) \textbf{Reflective Instruction}, which asks the model to analyze prior results before proposing a new prompt. In Stage~2, the meta-prompt additionally conditions on the discovered scaffold and requests small, task-specific refinements rather than rewriting the full prompt from scratch.
    }
    \label{fig:meta_prompt_example}
\end{figure*}

\noindent\textbf{Meta-Prompt:} a prompt provided to the Proposer LLM that specifies the optimization goal, the task format, and a history of previous candidate prompts and their scores. As shown in Figure~\ref{fig:meta_prompt_example}, this history guides the search toward improved instructions. In Stage 2, the same template additionally conditions on $P_{\text{scaffold}}$ and asks the Proposer to preserve it while making only limited task-specific refinements.

For fair comparison, all APO baselines, including OPRO, use the same Proposer/Scorer pair, the same meta-prompt template, and the same initial manual prompt $P_0$. The key difference is structural: baseline APO methods optimize each task-specific prompt directly from $P_0$, whereas DiSR first discovers a reusable scaffold and then adapts that scaffold to the target task.

\section{Experiments} 
\label{sec:experiments}

\subsection{Experimental Setup}
\label{sec:experimental_setup}

\paragraph{Datasets}
We evaluate DiSR on both commonsense and professional multiple-choice question answering benchmarks. For commonsense reasoning, we use LogiQA~\citep{liu2020logiqa}, OpenBookQA~\citep{Mihaylov2018CanAS}, and CosmosQA~\citep{huang2019cosmos}. For professional domains, we construct multi-task benchmarks in medicine, law, and finance using MedMCQA~\citep{pmlr-v174-pal22a}, AGIEval-Law~\citep{zhong2023agieval}, and FinEval~\citep{zhang2023fineval}. Within each domain, Scaffold Discovery uses source sub-tasks that share the same output format and evaluation metric but vary in topic or sub-type, while all remaining sub-tasks are held out for target-task evaluation. Detailed dataset statistics are provided in the appendix.

\paragraph{Models and Baselines}
DiSR uses PaLM 2-L-IT as the \textit{Proposer LLM} and PaLM 2-L as the \textit{Scorer LLM}, with 100 optimization iterations in each stage. The resulting prompts are evaluated on a set of external LLMs including GPT-4~\citep{openai2023gpt4}, GPT-3.5-Turbo~\citep{chatgpt}, LLaMA-2/3~\citep{touvron2023llama,llama3modelcard}, Vicuna~\citep{zheng2023judging}, and domain-specific models. We compare against a manual prompt and OPRO~\citep{yang2024largelanguagemodelsoptimizers}. For a controlled comparison, OPRO uses the same Proposer/Scorer pair, Stage~2 (target-task) optimization budget, initial manual prompt $P_0$, and meta-prompt template as DiSR, but directly optimizes a task-specific prompt from $P_0$ rather than first discovering a domain scaffold (Stage~1 is a one-time domain-level cost analyzed in Section~\ref{sec:opt_efficiency}). Additional APO baselines are reported in Appendix~\ref{sec:appendix_medmcqa_apo}.

\paragraph{Manual Initialization and Evaluation Metrics}
For all multiple-choice tasks, the shared manual starting prompt $P_0$ explicitly specifies the answer format:
\begin{quote}
\small
``The following are multiple-choice questions about professional [DOMAIN]. Please choose the correct answer from `A', `B', `C', or `D'.''
\end{quote}
This prompt is used both as the manual baseline and as the common initialization for OPRO and DiSR.

We report Accuracy (ACC), Expected Calibration Error (ECE)~\citep{guo2017calibration}, and Area Under the ROC Curve (AUROC)~\citep{fawcett2006introduction}, all in the zero-shot setting unless otherwise noted. Confidence scores for ECE and AUROC are derived from the model's answer probabilities following prior work~\citep{yang2023improving}. We interpret ECE narrowly as a measure of confidence calibration rather than as a broad notion of reliability. To analyze the decomposition hypothesis, we additionally study prompt semantics by encoding optimized prompts with all-mpnet-base-v2~\citep{reimers-2019-sentence-bert} and examining their relationships through t-SNE visualization~\citep{t-SNE}.

\begin{table*}[t]
\centering
\caption{
Performance comparison on commonsense reasoning benchmarks. We report mean scores over 3 random trials. DiSR is evaluated against a manual prompt and the OPRO baseline. Across most settings, DiSR improves over the manual baseline and often provides stronger accuracy and calibration than OPRO. Lower ECE indicates better confidence calibration. Best results are in \textbf{bold}.
}
\label{tab:main_results_definitive_final}

\fontsize{8.0}{9.0}\selectfont
\setlength{\tabcolsep}{6.0pt}
\renewcommand{\arraystretch}{1.06}

\begin{adjustbox}{max width=0.88\textwidth,center}
\begin{tabular}{@{}llccccccccc@{}}
\toprule
\textbf{Model} & \textbf{Meth.}
& \multicolumn{3}{c}{\textbf{LogiQA}}
& \multicolumn{3}{c}{\textbf{OpenBookQA}}
& \multicolumn{3}{c}{\textbf{CosmosQA}} \\
&
& \textbf{ACC} & \textbf{ECE} & \textbf{AUROC}
& \textbf{ACC} & \textbf{ECE} & \textbf{AUROC}
& \textbf{ACC} & \textbf{ECE} & \textbf{AUROC} \\
\midrule

LLaMA-2-7B & Manual
& 0.32 & 0.54 & 0.38
& 0.35 & 0.47 & 0.43
& 0.33 & 0.58 & 0.42 \\
& OPRO
& 0.31 & 0.48 & 0.45
& 0.37 & 0.41 & 0.48
& 0.35 & 0.42 & 0.49 \\
& \textbf{DiSR}
& \textbf{0.35} & \textbf{0.41} & \textbf{0.53}
& \textbf{0.38} & \textbf{0.37} & \textbf{0.55}
& \textbf{0.39} & \textbf{0.41} & \textbf{0.51} \\
\midrule

LLaMA-2-13B & Manual
& 0.34 & 0.47 & 0.47
& 0.39 & 0.46 & 0.45
& 0.41 & 0.46 & 0.59 \\
& OPRO
& \textbf{0.37} & 0.39 & 0.50
& 0.42 & 0.32 & 0.53
& 0.42 & 0.33 & 0.57 \\
& \textbf{DiSR}
& \textbf{0.37} & \textbf{0.31} & \textbf{0.57}
& \textbf{0.45} & \textbf{0.27} & \textbf{0.56}
& \textbf{0.45} & \textbf{0.30} & \textbf{0.62} \\
\midrule

LLaMA-3-8B & Manual
& 0.38 & 0.42 & 0.53
& 0.43 & 0.35 & 0.63
& 0.46 & 0.26 & 0.67 \\
& OPRO
& 0.41 & 0.36 & 0.61
& 0.51 & 0.26 & 0.71
& 0.49 & 0.20 & 0.68 \\
& \textbf{DiSR}
& \textbf{0.46} & \textbf{0.31} & \textbf{0.67}
& \textbf{0.55} & \textbf{0.21} & \textbf{0.75}
& \textbf{0.52} & \textbf{0.15} & \textbf{0.71} \\
\midrule

Vicuna-7B & Manual
& 0.29 & 0.64 & 0.44
& 0.32 & 0.49 & 0.37
& 0.31 & 0.51 & 0.47 \\
& OPRO
& 0.31 & 0.52 & \textbf{0.45}
& \textbf{0.36} & 0.45 & 0.44
& 0.33 & \textbf{0.46} & 0.58 \\
& \textbf{DiSR}
& \textbf{0.34} & \textbf{0.49} & 0.41
& \textbf{0.36} & \textbf{0.41} & \textbf{0.51}
& \textbf{0.34} & 0.53 & \textbf{0.63} \\
\midrule

Vicuna-13B & Manual
& 0.32 & 0.49 & 0.45
& 0.35 & 0.48 & 0.49
& 0.36 & 0.49 & 0.55 \\
& OPRO
& 0.35 & 0.42 & 0.51
& 0.40 & 0.43 & 0.54
& 0.40 & 0.39 & 0.61 \\
& \textbf{DiSR}
& \textbf{0.40} & \textbf{0.36} & \textbf{0.55}
& \textbf{0.44} & \textbf{0.39} & \textbf{0.58}
& \textbf{0.42} & \textbf{0.33} & \textbf{0.64} \\
\midrule

GPT-3.5-Turbo & Manual
& 0.35 & 0.42 & 0.61
& 0.37 & 0.36 & 0.58
& 0.42 & 0.39 & 0.61 \\
& OPRO
& 0.38 & 0.32 & 0.69
& 0.46 & 0.28 & 0.66
& 0.47 & 0.26 & 0.65 \\
& \textbf{DiSR}
& \textbf{0.42} & \textbf{0.26} & \textbf{0.73}
& \textbf{0.50} & \textbf{0.23} & \textbf{0.70}
& \textbf{0.51} & \textbf{0.20} & \textbf{0.68} \\
\midrule

GPT-4 & Manual
& 0.44 & 0.30 & 0.69
& 0.45 & 0.28 & 0.75
& 0.54 & 0.22 & 0.64 \\
& OPRO
& 0.48 & 0.23 & 0.78
& 0.54 & 0.21 & 0.78
& 0.56 & 0.19 & 0.71 \\
& \textbf{DiSR}
& \textbf{0.53} & \textbf{0.18} & \textbf{0.82}
& \textbf{0.59} & \textbf{0.16} & \textbf{0.83}
& \textbf{0.60} & \textbf{0.17} & \textbf{0.75} \\
\bottomrule
\end{tabular}
\end{adjustbox}
\end{table*}

\subsection{Main Performance Comparison}
\label{sec:results}

\begin{figure*}[t]
    \centering
    \includegraphics[width=.96\textwidth]{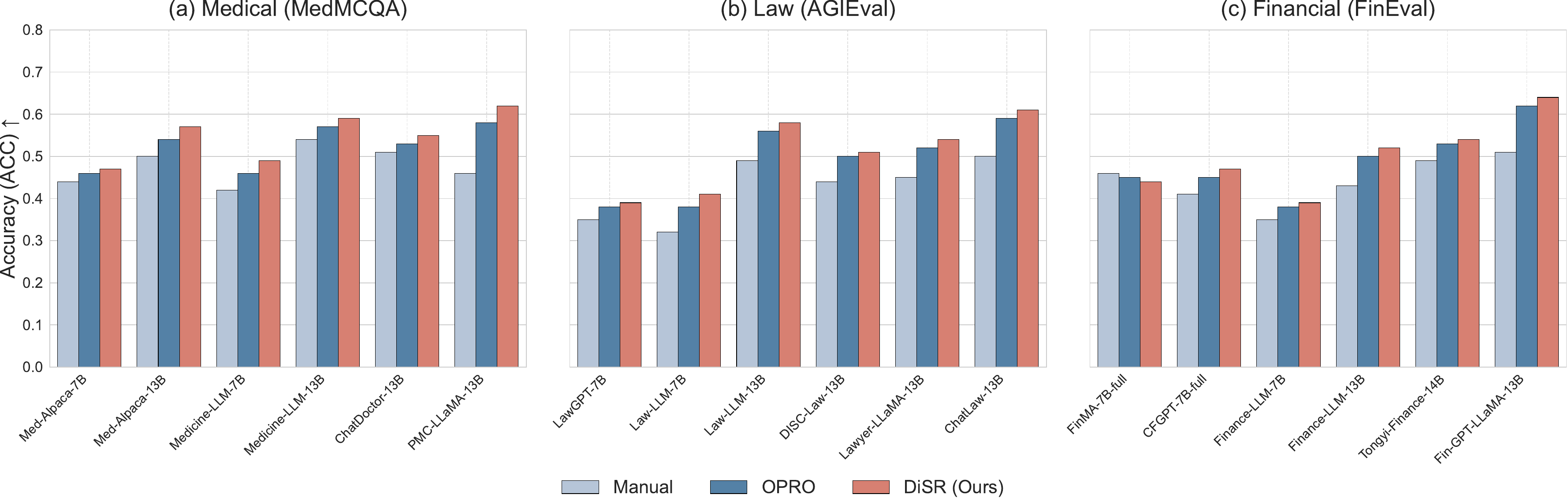}
    \caption{
        Performance on professional-domain benchmarks. Each subplot reports zero-shot accuracy (ACC) for medicine (MedMCQA), law (AGIEval-Law), and finance (FinEval). Across most settings, DiSR matches or outperforms the manual and OPRO baselines, suggesting that scaffold-based prompt reuse is particularly useful when related tasks share stable domain framing and output constraints.
    }
    \label{fig:domain_charts}
\end{figure*}

We first compare \textbf{DiSR} against a strong monolithic APO baseline, OPRO, on commonsense reasoning and professional-domain benchmarks. Table~\ref{tab:main_results_definitive_final} and Figure~\ref{fig:domain_charts} show a consistent overall pattern: across most evaluated models and tasks, DiSR improves over the manual prompt baseline and often outperforms OPRO in both accuracy and calibration. These gains are not uniform in every setting, but they suggest that explicitly separating a reusable scaffold from task-specific adaptation can provide a stronger starting point than optimizing each prompt independently from the same manual initialization.

On commonsense reasoning benchmarks (Table~\ref{tab:main_results_definitive_final}), DiSR is generally most effective on stronger evaluation models. For example, on GPT-4, DiSR achieves the best results across all three datasets, reaching 0.53/0.18/0.82 on LogiQA, 0.59/0.16/0.83 on OpenBookQA, and 0.60/0.17/0.75 on CosmosQA in terms of ACC/ECE/AUROC. Similar patterns are visible for GPT-3.5-Turbo and LLaMA-3-8B, where DiSR consistently improves both task performance and confidence calibration relative to OPRO, whereas on smaller or less capable models the advantage is narrower and occasionally mixed. Taken together, these results suggest that scaffold-based reuse is most beneficial when the downstream model can better exploit the learned instruction structure.

Results on professional domains (Figure~\ref{fig:domain_charts}) show the same overall tendency. Across medicine, law, and finance, DiSR typically matches or exceeds the manual and OPRO baselines, supporting the view that scaffold discovery can be useful in domains where related tasks share stable output constraints and domain-specific framing. We interpret these results as evidence of improved prompt reuse and target-task adaptation within coherent task families, rather than as evidence of dominance across all settings. We also compare against additional APO baselines, including Iterative-APE~\citep{zhou2022large}, PromptAgent~\citep{wang2023promptagent}, and APO~\citep{pryzant2023automatic}, on the medical benchmark (Appendix~\ref{sec:appendix_medmcqa_apo}). These results indicate that the gains of DiSR are not limited to comparison with a single monolithic baseline, although OPRO remains our primary controlled reference because it shares the same optimization backbone and differs mainly in prompt structure.

\begin{figure}[t]
    \centering
    \begin{subfigure}[t]{0.48\columnwidth}
        \centering
        \includegraphics[width=\columnwidth,height=4.2cm,keepaspectratio]{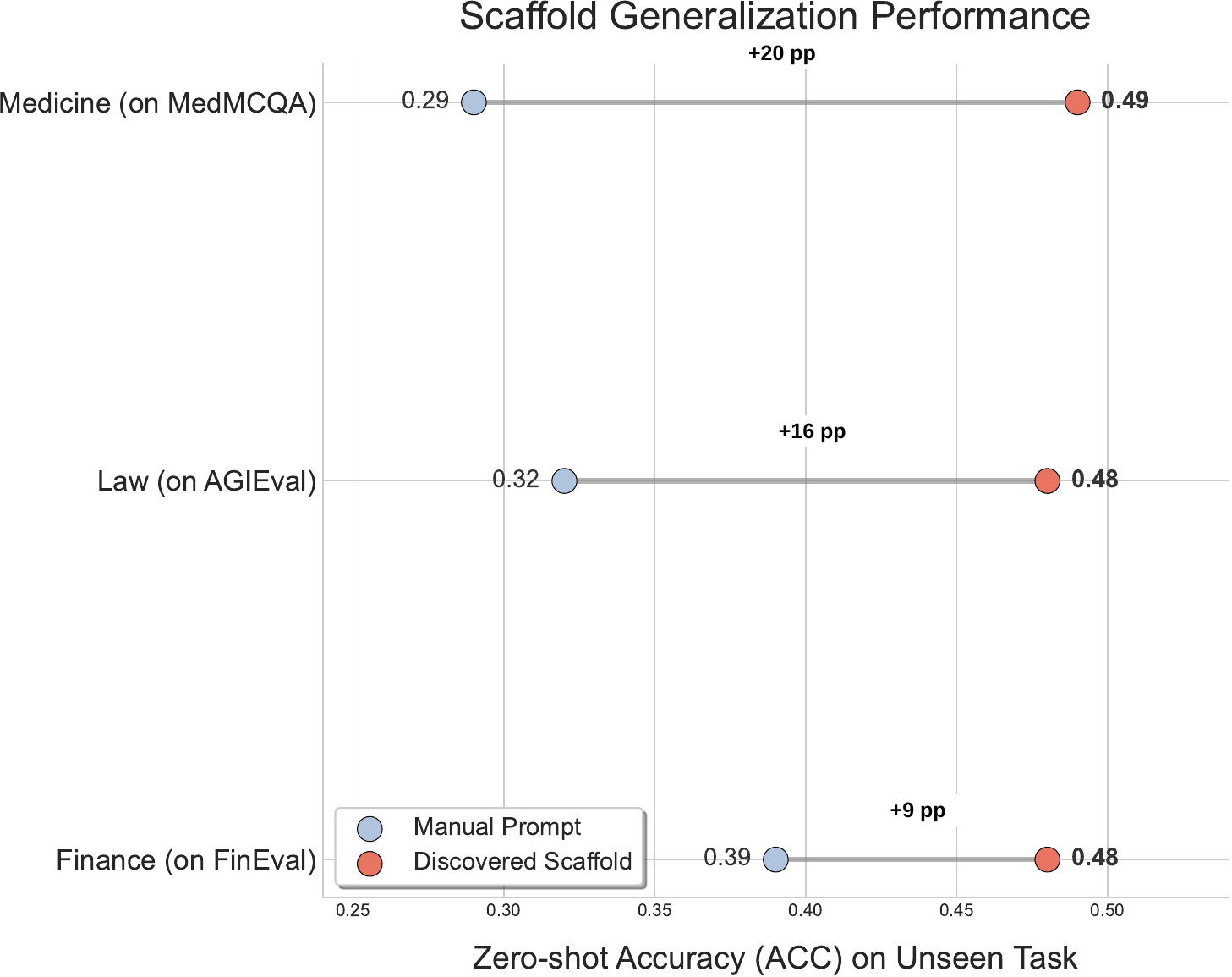}
        \caption{Scaffold Generalization}
        \label{fig:scaffold_generalization}
    \end{subfigure}\hfill
    \begin{subfigure}[t]{0.48\columnwidth}
        \centering
        \includegraphics[width=\columnwidth,height=4.2cm,keepaspectratio]{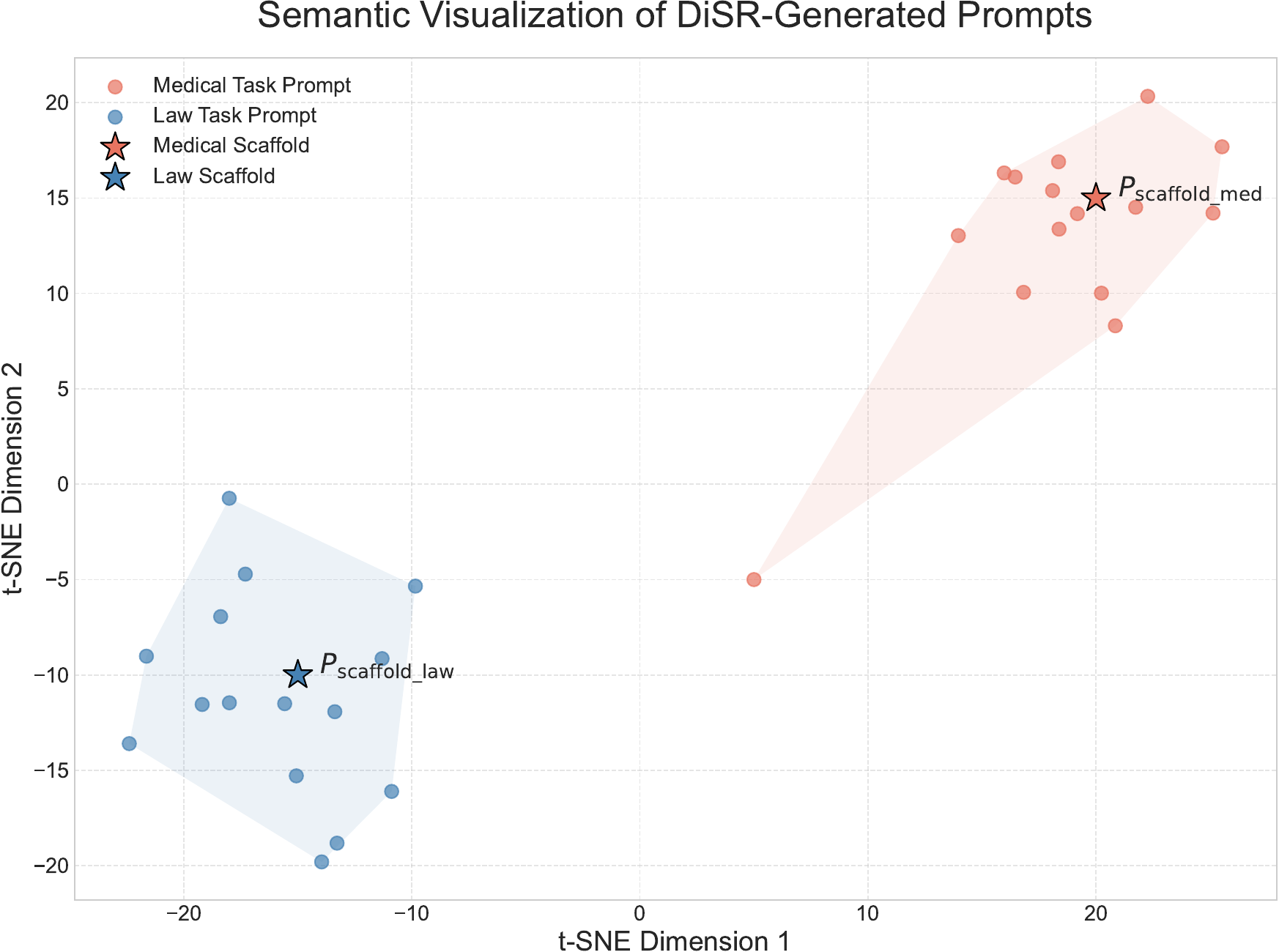}
        \caption{t-SNE of Prompt Space}
        \label{fig:tsne_visualization}
    \end{subfigure}
    \caption{
        Analyses of the discovered scaffold.
        \textbf{(a)} Zero-shot scaffold transfer to held-out tasks.
        \textbf{(b)} t-SNE visualization of optimized prompts and scaffolds.
    }
    \label{fig:scaffold_analysis}
\end{figure}

\subsection{Empirical Analysis of the Scaffold--Residual Decomposition}
Our central question is whether prompts for related tasks exhibit reusable structure that can be separated from task-specific adaptation. We first examine what the discovered scaffold captures, then study the role of the residual, and finally analyze the optimization efficiency enabled by scaffold reuse.

\subsubsection{Understanding the Scaffold}
A useful scaffold should transfer beyond the source tasks from which it was discovered. To test this, we perform a leave-one-out evaluation within each domain. For each held-out sub-task, we run Scaffold Discovery on the remaining source sub-tasks and then apply the resulting $P_{\text{scaffold}}$ directly to the held-out task in a zero-shot manner, without task-specific residual learning. Figure~\ref{fig:scaffold_analysis}(a) shows that the discovered scaffolds consistently outperform the generic manual prompt used in our main experiments. For example, in the medical domain, the scaffold improves zero-shot accuracy on the held-out MedMCQA task from 0.29 to 0.49, while in the finance domain it yields an absolute gain of 0.09. These results suggest that Stage~1 captures reusable domain-level instruction patterns that remain useful beyond the source tasks used to discover the scaffold. At the same time, scaffold-only performance is not sufficient for full task specialization, so we interpret these results as evidence that scaffold discovery provides a stronger domain-level starting point than a generic manual instruction, while Stage~2 remains necessary for task-specific refinement.

We also examine whether prompts produced by DiSR exhibit an interpretable organization in embedding space. To do so, we collect discovered scaffolds and final prompts from two domains (Medicine and Law), encode them with a sentence-transformer model, and project them into two dimensions using t-SNE. Figure~\ref{fig:scaffold_analysis}(b) reveals two qualitative patterns. First, prompts from the same domain tend to cluster together, suggesting that domain-level information is reflected in the learned prompt representations. Second, within each domain, the discovered scaffold appears near the center of the corresponding prompt group, with task-specialized prompts distributed around it. This pattern is consistent with the intended role of the scaffold as a reusable prompt backbone and the residual as a localized task-specific adaptation. We treat this visualization as qualitative support rather than as definitive evidence about prompt geometry. Since t-SNE can be sensitive to projection choices, its main value here is to illustrate that prompts optimized from the same scaffold tend to remain semantically closer to one another than to prompts from other domains.

\subsubsection{Understanding the Residual}
We next examine whether the learned residuals are scaffold-specific and whether they remain small relative to the scaffold. If the learned residual is task-specific rather than a generic improvement term, it should depend on the scaffold with which it was learned. We test this with a \emph{transplant} experiment: for a target task, we combine a scaffold from one domain with a residual learned in another domain and evaluate the resulting hybrid prompt. As shown in Figure~\ref{fig:residual_efficiency_analysis}(a), these mismatched scaffold--residual combinations perform substantially worse than the correctly matched prompts. This suggests that the residual is not a domain-agnostic performance booster. Instead, it behaves as a scaffold-dependent adaptation that is most effective when applied to the scaffold from which it was derived. We view this result as supporting the interpretation that DiSR separates shared prompt structure from task-specific modification.

We also analyze whether the learned residuals are in fact concise and task-targeted. Quantitatively, we measure the size of the Stage~2 adaptation using token-level edit distance, i.e., the number of tokens added, deleted, or substituted relative to the scaffold. Table~\ref{tab:residual_size} shows that these edits are consistently small: across all evaluated domains, the average residual accounts for less than 20\% of the scaffold length. We complement this analysis with a qualitative case study from the medical domain. Starting from the same medical scaffold, DiSR produces different task-specialized prompts for (A) diagnosis from clinical history and (B) diagnosis from laboratory findings. The resulting edits introduce task-relevant phrases such as ``based on the following \textbf{patient's clinical history}'' versus ``based on the following \textbf{laboratory test results}.'' This pattern suggests that Stage~2 does not simply rewrite the scaffold arbitrarily; instead, it tends to introduce a small amount of task-specific information that helps distinguish closely related tasks within the same domain. Taken together, these analyses suggest that DiSR specializes prompts through relatively small, scaffold-dependent edits rather than through wholesale prompt rewriting. This supports the practical interpretation of the residual as a concise task-level adaptation layered on top of a reusable scaffold.

\subsubsection{Optimization Efficiency Through Scaffold Reuse}
\label{sec:opt_efficiency}
\begin{figure}[t]
    \centering
    \begin{subfigure}[t]{0.48\columnwidth}
        \centering
        \includegraphics[width=\columnwidth,height=4.2cm,keepaspectratio]{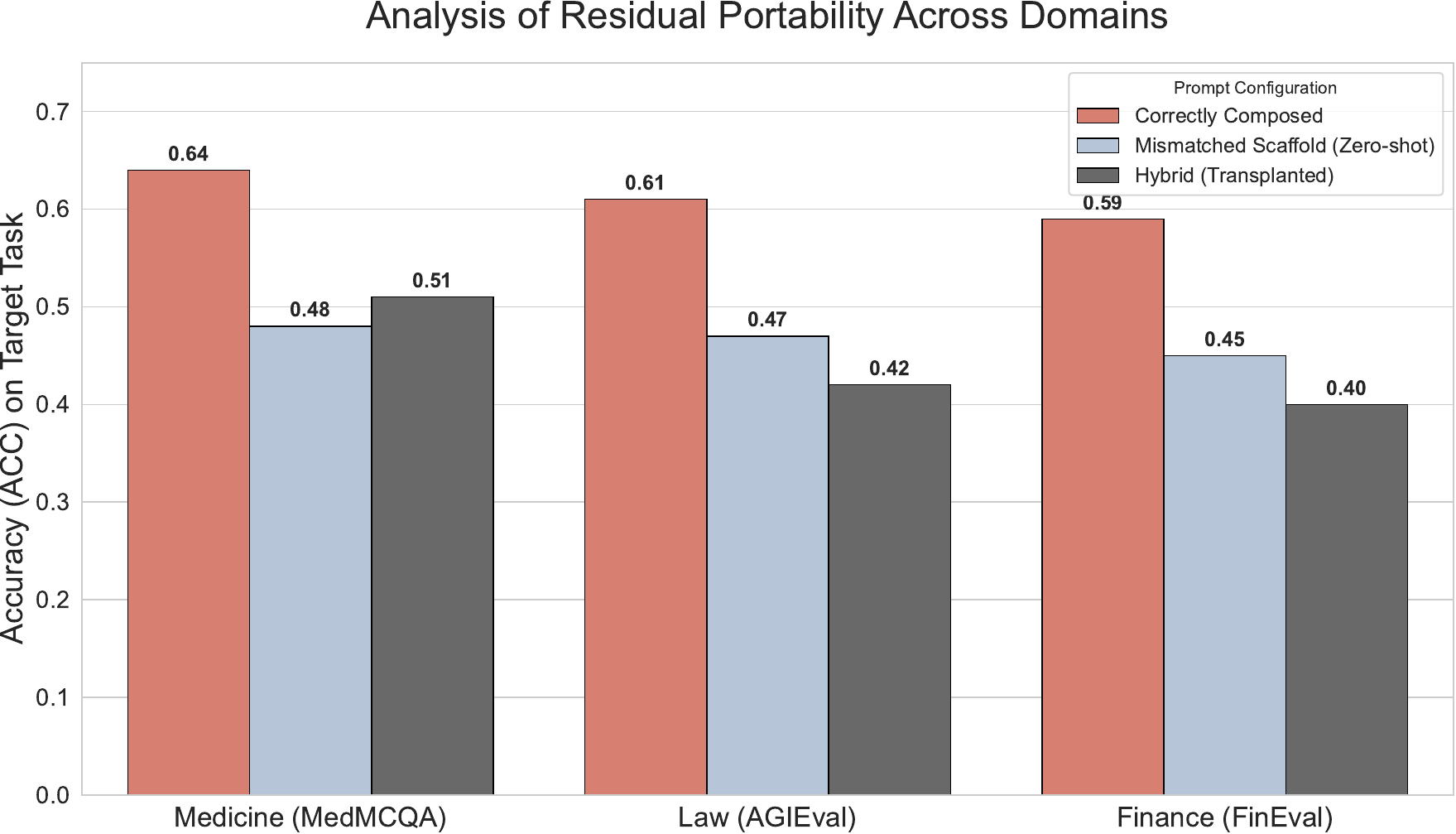}
        \caption{Residual Portability}
        \label{fig:residual_portability}
    \end{subfigure}\hfill
    \begin{subfigure}[t]{0.48\columnwidth}
        \centering
        \includegraphics[width=\columnwidth,keepaspectratio]{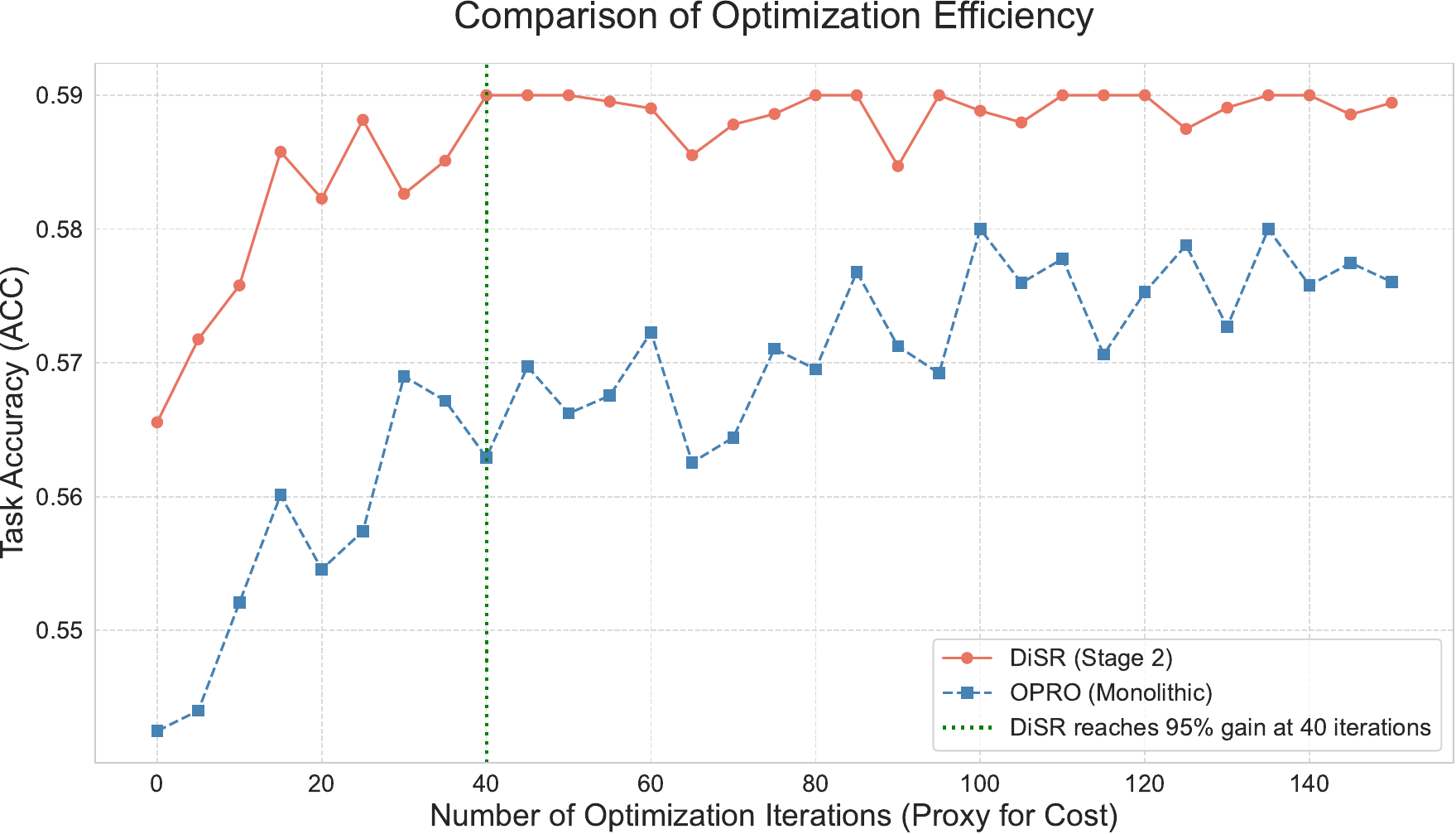}
        \caption{Optimization Efficiency}
        \label{fig:optimization_efficiency}
    \end{subfigure}
    \caption{
        Analyses of residual specificity and scaffold reuse.
        \textbf{(a)} Residual transplant across domains.
        \textbf{(b)} Stage~2 convergence compared with monolithic optimization.
    }
    \label{fig:residual_efficiency_analysis}
\end{figure}

A practical motivation for DiSR is that scaffold discovery is a one-time cost within a domain, after which target-task adaptation can be performed more efficiently. In our setup, Stage~1 runs for 100 iterations (evaluating each candidate across all $\kappa$ source tasks per iteration), whereas Stage~2 typically approaches its best performance in roughly 40 iterations. Figure~\ref{fig:residual_efficiency_analysis}(b) tracks convergence up to 150 iterations to verify stability beyond the default 100-iteration budget, showing that Stage~2 starts from a stronger initialization and converges faster than optimizing each target-task prompt from scratch with OPRO. This does not eliminate the upfront cost of scaffold discovery---which scales with $\kappa$---but it changes the cost structure: when multiple related target tasks are present within the same domain, the scaffold cost can be amortized across them. Under our experimental budget, this makes DiSR increasingly attractive as the number of target tasks grows. We therefore interpret the efficiency benefit as a domain-level reuse effect rather than as a universal reduction in optimization cost for every single task.

\begin{table}[htbp]
    \centering
    \caption{Component lengths for discovered scaffolds and learned residuals. Residuals remain substantially shorter than scaffolds across all domains.}
    \label{tab:residual_size}
    \setlength{\tabcolsep}{6pt}
    \renewcommand{\arraystretch}{1.2}
    \begin{adjustbox}{max width=0.93\columnwidth,center}
    \begin{tabular}{@{}lccc@{}}
        \toprule
        \rowcolor{gray!10}
        \textbf{Domain} & \textbf{Scaffold Length} & \textbf{Avg. Residual Length} & \textbf{Residual as \% of Scaffold} \\
        \rowcolor{gray!6}
         & \textbf{(tokens)} & \textbf{(tokens)} &  \\
        \midrule
        Medicine & 75 & 7.1  & \textbf{9.5\%} \\
        Law      & 88 & 15.4 & \textbf{17.5\%} \\
        Finance  & 52 & 9.8  & \textbf{18.8\%} \\
        \bottomrule
    \end{tabular}
    \end{adjustbox}
\end{table}

\section{Conclusion} \label{sec:conclusion} We studied a structured view of automatic prompt optimization in which prompts for related tasks are decomposed into a reusable \textbf{Instructional Scaffold} and a \textbf{Task-specific Residual}. Based on this view, we proposed \textbf{DiSR}, a two-stage framework that first discovers a scaffold from source tasks within a domain and then adapts it to a target task through a residual. Across commonsense and professional-domain benchmarks, our results show that this decomposition can improve prompt reuse and target-task adaptation relative to monolithic optimization from the same manual initialization, often with favorable calibration and optimization efficiency. Additional analyses of scaffold transfer, residual portability, and residual size further suggest that prompts within coherent task families can exhibit reusable structure together with compact task-specific variation. More broadly, our findings indicate that making shared and task-specific prompt content explicit can provide a useful empirical framework for studying prompt reuse across related NLP tasks.

\section*{Limitations}
\label{sec:limitations}

DiSR has several limitations. Its effectiveness depends on domain coherence: if source tasks are too heterogeneous in format or reasoning pattern, Stage~1 may fail to discover a useful scaffold. Its benefits are also model-dependent; in our experiments, DiSR is generally more effective on stronger models, while gains on smaller models are narrower and sometimes mixed. In addition, our empirical evidence is still concentrated on multiple-choice question answering, with only a small extension to SST-2, so it remains unclear how well the scaffold--residual decomposition transfers to broader settings such as open-ended generation, multi-step interaction, or agentic workflows. Finally, ECE should be interpreted narrowly as a measure of confidence calibration rather than as a broad notion of reliability, since we do not evaluate robustness, fairness, factuality, or resistance to adversarial prompting.

\section*{Acknowledgments}
This work is supported by the New Generation Artificial Intelligence--National Science and Technology Major Project (No.2026ZD0127902) and the Fundamental and Interdisciplinary Disciplines Breakthrough Plan of the Ministry of Education of China (JYB2025XDXM903).

\bibliography{custom,additional_refs,zhan_refs}

\clearpage

\appendix
\renewcommand{\rowcolor}[1]{}
\vspace{2em}
\begin{center}
    \Large{\textbf{Appendix}}
\end{center}
\vspace{2em}

\etocdepthtag.toc{appendix}
\etocsettagdepth{chapter}{none}
\etocsettagdepth{appendix}{subsection}
\tableofcontents

\section{Experimental Details}
\label{sec:appendix_experimental_details}

This appendix provides additional details about the experimental setup, including evaluation models, datasets, hyperparameters, and implementation details.

\subsection{Evaluation Models and Datasets}
\label{sec:appendix_models_datasets}

\noindent\textbf{Evaluation Models.}
We evaluate the prompts generated by \textbf{DiSR} and the baselines on a set of external large language models, including both general-purpose foundation models and models adapted to professional domains.
\begin{itemize}
    \item \textbf{Foundational Models (7):} \textit{GPT-4}~\citep{openai2023gpt4}, \textit{GPT-3.5-Turbo}~\citep{chatgpt}, \textit{LLaMA-3-8B}~\citep{llama3modelcard}, \textit{LLaMA-2-7B}, \textit{LLaMA-2-13B}~\citep{touvron2023llama}, \textit{Vicuna-7B}, and \textit{Vicuna-13B}~\citep{zheng2023judging}.
    \item \textbf{Domain-specific Models (18):} To examine behavior in professional settings, we include models adapted to medicine, law, and finance:
    \begin{itemize}
        \item \textit{Medicine (6):} \textit{ChatDoctor-13B}~\citep{li2023chatdoctor}, \textit{PMC-LLaMA-13B}~\citep{wu2023pmcllama}, \textit{MedAlpaca-7B}, \textit{MedAlpaca-13B}~\citep{han2023medalpaca}, \textit{Medicine-LLM-7B}, and \textit{Medicine-LLM-13B}~\citep{cheng2023adapting}.
        \item \textit{Law (6):} \textit{ChatLaw-13B}~\citep{cui2024chatlaw}, \textit{DISC-LawLLM-13B}~\citep{yue2023disclawllm}, \textit{Lawyer-LLaMA-13B}~\citep{huang2023lawyer}, \textit{LawGPT-7B}~\citep{zhou2024lawgpt}, \textit{Law-LLM-7B}, and \textit{Law-LLM-13B}~\citep{cheng2023adapting}.
        \item \textit{Finance (6):} \textit{FinGPT-13B-v2}~\citep{yang2023fingpt}, \textit{Tongyi-Finance-14B-Chat}, \textit{FinMA-7B-full}~\citep{xie2023pixiu}, \textit{CFGPT-7B-full}~\citep{li2023cfgpt}, \textit{Finance-LLM-7B}, and \textit{Finance-LLM-13B}~\citep{cheng2023adapting}.
    \end{itemize}
\end{itemize}

\noindent\textbf{Evaluation Datasets.}
Consistent with the main paper, we evaluate DiSR on both commonsense and professional multiple-choice question answering benchmarks.
\begin{itemize}
    \item \textit{Commonsense Reasoning:} We use LogiQA~\citep{liu2020logiqa}, OpenBookQA~\citep{Mihaylov2018CanAS}, and CosmosQA~\citep{huang2019cosmos}. Specifically, we evaluate on the official test sets of \textbf{LogiQA} (651 questions) and \textbf{OpenBookQA} (500 questions). For \textbf{CosmosQA}, we randomly sample 1,000 questions from the validation split.
    \item \textit{Professional Domains:} We construct multi-task benchmarks in medicine, law, and finance using \textbf{MedMCQA}~\citep{pmlr-v174-pal22a}, \textbf{AGIEval-Law}~\citep{zhong2023agieval}, and \textbf{FinEval}~\citep{zhang2023fineval}. Within each domain, Scaffold Discovery uses source sub-tasks that share the same output format and evaluation metric but vary in topic or sub-type, while the remaining sub-tasks are held out for target-task evaluation. For evaluation, we use 1,000 questions sampled from the MedMCQA test split, all questions from the \texttt{jec-qa-kd} and \texttt{jec-qa-ca} law tasks in AGIEval, and all questions from the validation splits of \texttt{advanced financial accounting}, \texttt{financial markets}, and \texttt{corporate finance} in FinEval.
\end{itemize}

\subsection{Hyperparameters and Implementation Details}
\label{sec:appendix_hyperparams_impl}

\noindent\textbf{Hyperparameters.}
The main hyperparameters of the optimization loop are listed in Table~\ref{tab:hyperparameters}. Consistent with the main paper, DiSR uses PaLM 2-L-IT as the Proposer LLM and PaLM 2-L as the Scorer LLM. We use temperature 0.7 and top-p 0.95 for the Proposer LLM, and each DiSR stage is run for 100 optimization iterations. The meta-prompt includes 10 historical prompt-score pairs as optimization context.

\begin{table}[htbp]
    \centering
    \caption{Core hyperparameters for the DiSR optimization loop.}
    \label{tab:hyperparameters}
    \fontsize{8.0}{9.0}\selectfont
    \setlength{\tabcolsep}{4.5pt}
    \renewcommand{\arraystretch}{1.06}
    \begin{tabular}{@{}ll@{}}
        \toprule
        \textbf{Hyperparameter} & \textbf{Value} \\
        \midrule
        \multicolumn{2}{@{}l@{}}{\textit{\textbf{Proposer LLM (PaLM 2-L-IT) Settings}}} \\
        Temperature & 0.7 \\
        Top-p & 0.95 \\
        Number of Candidates per Step & 8 \\
        \midrule
        \multicolumn{2}{@{}l@{}}{\textit{\textbf{Optimization Loop Settings}}} \\
        Iterations per Stage (Stages 1 \& 2) & 100 \\
        Historical Exemplars in Meta-Prompt & 10 \\
        \bottomrule
    \end{tabular}
\end{table}

\noindent\textbf{Meta-Prompt.}
For completeness, we provide the meta-prompt template (Figure~\ref{fig:appendix_meta_prompt_template}) and one instantiated example (Figure~\ref{fig:appendix_meta_prompt_instance}) below.

\begin{figure*}[t!]
    \centering
    \begin{tcolorbox}[colback=gray!5!white,colframe=gray!75!black,title=Meta-Prompt Template,fonttitle=\bfseries]
    \footnotesize
    \begin{verbatim}
{{OPTIMIZATION_HISTORY}}

The subsequent examples illustrate the application of your instruction:
replace <INS> in each input with your instruction, then read the input
and generate an output. An output is deemed correct if it matches the
provided output; otherwise, it is considered incorrect.

{{TASK_EXAMPLES}}

First, briefly analyze the provided instructions and their scores. What
makes higher-scoring prompts more effective? Based on this analysis,
propose a new, improved instruction that aims to achieve the highest
possible score. Enclose only the new instruction within square brackets.
    \end{verbatim}
    \end{tcolorbox}
    \caption{Template for the meta-prompt used in the optimization loop.}
    \label{fig:appendix_meta_prompt_template}
\end{figure*}

\begin{figure*}[t!]
    \centering
    \begin{tcolorbox}[colback=blue!5!white,colframe=blue!75!black,
        title=Example Filled-in Meta-Prompt (Medical Domain),fonttitle=\bfseries]
    \footnotesize
    \begin{verbatim}
Prompt1:
The following are multiple-choice questions (with answers) about
professional medicine. Please choose the correct answer from "A", "B",
"C", or "D".
Score: 61

Prompt2:
Presented below is a multiple-choice question pertaining to professional
medicine. Carefully review the question and select the correct option
from "A", "B", "C", or "D".
Score: 63

The subsequent examples illustrate the application of your instruction:
replace <INS> in each input with your instruction, then read the input
and generate an output. An output is deemed correct if it matches the
provided output; otherwise, it is considered incorrect.

Input:
A: <INS>
Q: A 13-month-old child is brought to the emergency department because
of urticaria, swelling of the lips, and difficulty breathing
immediately after eating an egg. A potential risk for hypersensitivity
reaction is posed by vaccination against which of the following
illnesses? A: Hepatitis B: Influenza C: Pertussis D: Poliomyelitis
Output: B

First, briefly analyze the provided instructions and their scores. What
makes higher-scoring prompts more effective? Based on this analysis,
propose a new, improved instruction that aims to achieve the highest
possible score. Enclose only the new instruction within square brackets.
    \end{verbatim}
    \end{tcolorbox}
    \caption{An instantiated meta-prompt example for a medical optimization step.}
    \label{fig:appendix_meta_prompt_instance}
\end{figure*}

\section{Additional Results and Analysis}
\label{sec:appendix_additional_results}

This appendix provides additional quantitative results and qualitative examples that complement the analyses in the main paper.

\subsection{Performance in the Few-shot Setting}
\label{sec:appendix_fewshot}

To examine whether prompts discovered by DiSR remain useful in in-context learning, we additionally evaluate them in a five-shot setting on the commonsense reasoning benchmarks. Table~\ref{tab:fewshot_results_commonsense} shows that the main pattern from the zero-shot experiments largely remains: DiSR generally improves over the manual prompt baseline and often outperforms OPRO on accuracy, while calibration trends are broadly similar to those reported in the main results. In a small number of cases, OPRO remains competitive on secondary metrics, but the overall picture suggests that the scaffold--residual decomposition continues to provide a useful prompt structure when few-shot demonstrations are added.

\begin{table*}[t]
\centering
\caption{
Five-shot performance comparison on commonsense reasoning benchmarks. We report mean scores over 3 random trials. Across most settings, DiSR improves over the manual baseline and often outperforms OPRO. Best performance for each metric is in \textbf{bold}.
}
\label{tab:fewshot_results_commonsense}

\fontsize{8.0}{9.0}\selectfont
\setlength{\tabcolsep}{6.25pt}
\renewcommand{\arraystretch}{1.06}

\begin{adjustbox}{max width=0.88\textwidth,center}
\begin{tabular}{@{}l@{\hskip 2pt}lccccccccc@{}}
\toprule
\multicolumn{2}{c}{}
& \multicolumn{3}{c}{\textbf{LogiQA}}
& \multicolumn{3}{c}{\textbf{OpenBookQA}}
& \multicolumn{3}{c}{\textbf{CosmosQA}} \\
\textbf{Model} & \textbf{Meth.}
& \textbf{ACC} & \textbf{ECE} & \textbf{AUROC}
& \textbf{ACC} & \textbf{ECE} & \textbf{AUROC}
& \textbf{ACC} & \textbf{ECE} & \textbf{AUROC} \\
\midrule

\multirow{3}{*}{LLaMA-2-7B}
& Manual
& 0.37 & 0.52 & 0.45
& 0.40 & 0.49 & 0.55
& 0.37 & 0.52 & 0.45 \\
& OPRO
& 0.40 & 0.44 & 0.51
& 0.41 & 0.46 & 0.57
& 0.40 & 0.49 & 0.52 \\
& \textbf{DiSR}
& \textbf{0.41} & \textbf{0.39} & \textbf{0.55}
& \textbf{0.44} & \textbf{0.42} & \textbf{0.61}
& \textbf{0.43} & \textbf{0.45} & \textbf{0.56} \\
\midrule

\multirow{3}{*}{LLaMA-2-13B}
& Manual
& 0.38 & 0.45 & 0.49
& 0.37 & 0.52 & 0.41
& 0.46 & 0.47 & 0.63 \\
& OPRO
& 0.41 & 0.40 & 0.55
& 0.47 & 0.34 & \textbf{0.59}
& 0.45 & 0.38 & 0.62 \\
& \textbf{DiSR}
& \textbf{0.43} & \textbf{0.37} & \textbf{0.57}
& \textbf{0.49} & \textbf{0.31} & \textbf{0.59}
& \textbf{0.48} & \textbf{0.34} & \textbf{0.67} \\
\midrule

\multirow{3}{*}{LLaMA-3-8B}
& Manual
& 0.43 & 0.35 & 0.67
& 0.44 & 0.30 & 0.67
& 0.46 & 0.25 & 0.67 \\
& OPRO
& 0.47 & 0.26 & 0.77
& 0.55 & 0.20 & 0.81
& 0.52 & \textbf{0.16} & 0.75 \\
& \textbf{DiSR}
& \textbf{0.50} & \textbf{0.24} & \textbf{0.79}
& \textbf{0.58} & \textbf{0.17} & \textbf{0.84}
& \textbf{0.56} & 0.18 & \textbf{0.79} \\
\midrule

\multirow{3}{*}{GPT-4}
& Manual
& 0.47 & 0.26 & 0.75
& 0.50 & 0.21 & 0.78
& 0.55 & 0.18 & 0.70 \\
& OPRO
& 0.54 & 0.17 & 0.85
& 0.61 & 0.15 & 0.84
& 0.63 & 0.11 & 0.86 \\
& \textbf{DiSR}
& \textbf{0.56} & \textbf{0.14} & \textbf{0.89}
& \textbf{0.63} & \textbf{0.12} & \textbf{0.87}
& \textbf{0.65} & \textbf{0.10} & \textbf{0.88} \\
\bottomrule
\end{tabular}
\end{adjustbox}
\end{table*}

\subsection{Qualitative Examples and Analysis}
\label{sec:appendix_qualitative_analysis}

\noindent\textbf{Qualitative Examples of Discovered Scaffolds and Task-specific Edits.}
To make the decomposition more concrete, we present one qualitative example from each domain (Tables~\ref{tab:qualitative_example_commonsense}--\ref{tab:qualitative_example_fin}). Across these examples, a recurring pattern is that the \textbf{Instructional Scaffold} specifies the broad role, output format, and general reasoning frame, while the final prompt adds a relatively small amount of task-specific content.

\begin{table*}[t!]
    \centering
    \small
    \caption{Qualitative example from the \textbf{Commonsense Reasoning} domain (LogiQA).}
    \label{tab:qualitative_example_commonsense}
    \setlength{\tabcolsep}{2.0pt}
    \renewcommand{\arraystretch}{1.06}
    \begin{tabular}{@{}p{0.23\textwidth} p{0.73\textwidth}@{}}
        \toprule
        \textbf{Component} & \textbf{Generated Prompt Text} \\
        \midrule
        \textbf{Instructional Scaffold} ($P_{\text{scaffold\_commonsense}}$)
        & Your task is to solve the following reasoning puzzle. Analyze the provided text and question carefully. Select the most logical conclusion from the options ``A'', ``B'', ``C'', or ``D''. \\
        \textbf{Final Prompt} ($P_k^*$) for a deductive reasoning task
        & Your task is to solve the following reasoning puzzle. Analyze the provided text and question carefully. \textbf{Identify the key premises and evaluate the validity of the logical chain of inference.} Select the most logical conclusion from the options ``A'', ``B'', ``C'', or ``D''. \\
        \bottomrule
    \end{tabular}
\end{table*}

\begin{table*}[t!]
    \centering
    \small
    \caption{Qualitative example from the \textbf{Medical} domain.}
    \label{tab:qualitative_example_med}
    \setlength{\tabcolsep}{2.0pt}
    \renewcommand{\arraystretch}{1.06}
    \begin{tabular}{@{}p{0.23\textwidth} p{0.73\textwidth}@{}}
        \toprule
        \textbf{Component} & \textbf{Generated Prompt Text} \\
        \midrule
        \textbf{Instructional Scaffold} ($P_{\text{scaffold\_med}}$)
        & As an expert medical professional, analyze the following clinical scenario. Your task is to select the single best answer from the options ``A'', ``B'', ``C'', or ``D''. Your decision should be based on established medical guidelines and evidence-based practice. \\
        \textbf{Final Prompt} ($P_k^*$) for ``Diagnosis from History''
        & As an expert medical professional, analyze the following clinical scenario. Your task is to select the single best answer from the options ``A'', ``B'', ``C'', or ``D''. \textbf{Focus specifically on the patient's clinical history and symptoms to determine the most likely diagnosis.} Your decision should be based on established medical guidelines and evidence-based practice. \\
        \bottomrule
    \end{tabular}
\end{table*}

\begin{table*}[t!]
    \centering
    \small
    \caption{Qualitative example from the \textbf{Law} domain.}
    \label{tab:qualitative_example_law}
    \setlength{\tabcolsep}{2.0pt}
    \renewcommand{\arraystretch}{1.06}
    \begin{tabular}{@{}p{0.23\textwidth} p{0.73\textwidth}@{}}
        \toprule
        \textbf{Component} & \textbf{Generated Prompt Text} \\
        \midrule
        \textbf{Instructional Scaffold} ($P_{\text{scaffold\_law}}$)
        & As a legal analyst, you are presented with a case. Your objective is to identify the most accurate legal conclusion from the choices ``A'', ``B'', ``C'', or ``D''. Your analysis must be grounded in relevant statutes and legal precedents. \\
        \textbf{Final Prompt} ($P_k^*$) for ``Contract Dispute''
        & As a legal analyst, you are presented with a case. Your objective is to identify the most accurate legal conclusion from the choices ``A'', ``B'', ``C'', or ``D''. \textbf{Evaluate the formation and validity of the contract in question, paying close attention to the elements of offer, acceptance, and consideration.} Your analysis must be grounded in relevant statutes and legal precedents. \\
        \bottomrule
    \end{tabular}
\end{table*}

\begin{table*}[t!]
    \centering
    \small
    \caption{Qualitative example from the \textbf{Finance} domain.}
    \label{tab:qualitative_example_fin}
    \setlength{\tabcolsep}{2.0pt}
    \renewcommand{\arraystretch}{1.06}
    \begin{tabular}{@{}p{0.23\textwidth} p{0.73\textwidth}@{}}
        \toprule
        \textbf{Component} & \textbf{Generated Prompt Text} \\
        \midrule
        \textbf{Instructional Scaffold} ($P_{\text{scaffold\_fin}}$)
        & You are a financial analyst. Based on the data provided, select the best financial recommendation from options ``A'', ``B'', ``C'', or ``D''. Your answer should align with standard principles of financial analysis and risk management. \\
        \textbf{Final Prompt} ($P_k^*$) for ``Equity Valuation''
        & You are a financial analyst. Based on the data provided, select the best financial recommendation from options ``A'', ``B'', ``C'', or ``D''. \textbf{Your primary task is to perform an equity valuation; consider metrics such as the P/E ratio, discounted cash flow (DCF), and market comparables.} Your answer should align with standard principles of financial analysis and risk management. \\
        \bottomrule
    \end{tabular}
\end{table*}

\noindent\textbf{Analysis of Lower-scoring Prompts.}
Inspecting lower-scoring prompts also helps clarify what Stage~1 appears to capture. For example, an early prompt such as \textit{``Just answer the question.''} is too underspecified to provide useful task framing, while a prompt such as \textit{``As a legal expert specializing in maritime law, answer \ldots''} is overly narrow for a broader legal domain. These cases are consistent with the intuition that an effective scaffold should be specific enough to provide useful guidance while remaining broad enough to transfer across related tasks within the same domain.

\subsection{Additional APO Baselines on MedMCQA}
\label{sec:appendix_medmcqa_apo}

To supplement the main comparison with OPRO, we also evaluate DiSR against several additional APO baselines on MedMCQA: Iterative-APE~\citep{zhou2022large}, PromptAgent~\citep{wang2023promptagent}, and APO~\citep{pryzant2023automatic}. These methods use the same Proposer/Scorer LLM pair and the same meta-prompt template described in Section~\ref{sec:experimental_setup}, but differ in search strategy. Unless otherwise noted, we report train and test accuracy on MedMCQA using Medicine-LLM-13B~\citep{cheng2023adapting} as the evaluation model.

\begin{table}[htbp]
    \centering
    \caption{Average train and test accuracy on MedMCQA for DiSR and additional APO baselines (evaluated with Medicine-LLM-13B).}
    \label{tab:medmcqa_train_test}
    \fontsize{8.0}{9.0}\selectfont
    \setlength{\tabcolsep}{4.5pt}
    \renewcommand{\arraystretch}{1.06}
    \begin{tabular}{@{}lcc@{}}
        \toprule
        \textbf{Method} & \textbf{Train ACC} & \textbf{Test ACC} \\
        \midrule
        \textbf{DiSR (Ours)} & \textbf{0.68} & \textbf{0.63} \\
        OPRO                 & 0.60          & 0.55          \\
        PromptAgent          & 0.59          & 0.53          \\
        Iterative-APE        & 0.57          & 0.52          \\
        APO                  & 0.54          & 0.50          \\
        \bottomrule
    \end{tabular}
\end{table}

Table~\ref{tab:medmcqa_train_test} shows that DiSR achieves the highest train and test accuracy among the compared methods in this setting. We interpret this result as supplementary evidence that the gains observed for DiSR are not limited to comparison with OPRO alone. At the same time, because these methods differ in search strategy as well as prompt parameterization, we treat OPRO as the primary controlled baseline in the main paper.

\subsection{A Non-MCQA Task: Sentiment Classification on SST-2}
\label{sec:appendix_sst2}

To examine whether the scaffold--residual decomposition extends beyond multiple-choice QA, we conduct a small-scale experiment on SST-2. We cast sentiment classification as a two-label prediction task and apply the same overall DiSR procedure: Stage~1 discovers a sentiment-analysis scaffold from source subsets, and Stage~2 adapts it to the final evaluation split. Prompts are evaluated in a zero-shot setting, and we report Accuracy (ACC) and Expected Calibration Error (ECE).

\begin{table}[htbp]
    \centering
    \caption{Performance on SST-2 sentiment classification (zero-shot). DiSR improves over the manual prompt and OPRO in this small non-MCQA experiment.}
    \label{tab:sst2_results}
    \fontsize{8.0}{9.0}\selectfont
    \setlength{\tabcolsep}{4.5pt}
    \renewcommand{\arraystretch}{1.06}
    \begin{tabular}{@{}llcc@{}}
        \toprule
        \textbf{Model} & \textbf{Method} & \textbf{ACC} ($\uparrow$) & \textbf{ECE} ($\downarrow$) \\
        \midrule
        \multirow{3}{*}{GPT-3.5-Turbo}
          & Manual        & 0.80 & 0.32 \\
          & OPRO          & 0.87 & 0.24 \\
          & \textbf{DiSR (Ours)} & \textbf{0.91} & \textbf{0.16} \\
        \midrule
        \multirow{3}{*}{LLaMA-2-13B}
          & Manual        & 0.71 & 0.38 \\
          & OPRO          & 0.78 & 0.34 \\
          & \textbf{DiSR (Ours)} & \textbf{0.82} & \textbf{0.22} \\
        \bottomrule
    \end{tabular}
\end{table}

On SST-2 (Table~\ref{tab:sst2_results}), DiSR follows the same overall pattern observed on the MCQA tasks: for both GPT-3.5-Turbo and LLaMA-2-13B, it improves over the manual prompt and OPRO on both accuracy and ECE. We present this experiment as a small extension beyond the main task setting rather than as definitive evidence of broad task generality. Extending prompt decomposition to more complex settings---such as multi-step context allocation and agentic search~\citep{liu2026laws,he2026localsearchbench}, multi-turn tool-use~\citep{chen2026toolforge}, or hierarchical calibration in structured code execution~\citep{liu2026neurosym}---remains for future work. Further settings for investigation include thematic analysis~\citep{qiao2025thematic} and prompt-driven zero-shot domain generalization~\citep{qiao2026zero}. Even so, the results suggest that the scaffold--residual decomposition may remain useful outside the MCQA setting when related classification tasks share a common decision format. Further work could examine sensitivity to language roles in multilingual tasks~\citep{zhan2026disentangling} and to instruction wording~\citep{zhan2026verifiable}.

\section{More Discussions}

\subsection{Reproducibility Statement}
\label{sec:reproducibility}

We provide the main information needed to reproduce our experimental setup. All datasets are drawn from public benchmarks, and the evaluation models and data splits are listed in Appendix~\ref{sec:appendix_models_datasets}. The optimization framework is described in Section~\ref{sec:methods}, and Appendix~\ref{sec:appendix_hyperparams_impl} reports the main hyperparameters together with the meta-prompt template used in the optimization loop. We also specify the shared manual initialization, the comparison protocol with OPRO, and the evaluation metrics used throughout the paper.

At the same time, exact reproduction may depend on access to the same proprietary optimizer models and APIs used in our experiments. Our goal is therefore to make the setup as transparent as possible so that the main comparisons and design choices can be independently verified and extended. Assessing the stability of method rankings under limited evaluation data, as studied in Bayesian deep learning~\citep{zhan2026unstable}, is another direction for future evaluation.

\subsection{Ethics Statement}
This work uses public benchmark datasets and evaluates prompt optimization methods in controlled experimental settings. It does not involve private user data or human-subject annotation. Because several evaluated benchmarks concern professional domains such as medicine, law, and finance, the results should not be interpreted as evidence that the resulting prompts are suitable for real-world decision-making without additional safeguards and domain-specific validation. Moreover, because prompt structure and continuation framing can influence in-context behavioral alignment~\citep{liu2026harmful}, any practical deployment of automatically optimized prompts should also incorporate framing-sensitive safety evaluation.

\subsection{LLM Usage Statement}
Large language models were used during manuscript preparation for language assistance and editing. All authors reviewed the final manuscript and take responsibility for its content.

\end{document}